\documentclass{article}
\usepackage[T1]{fontenc}
\usepackage{microtype}
\usepackage{graphicx}
\usepackage{dblfloatfix}
\usepackage{caption}
\usepackage{subcaption}
\usepackage{booktabs} % for professional tables
\usepackage{amsmath}
\usepackage{amssymb}
\usepackage{tikz}

\usepackage{hyperref}

\usepackage[accepted]{icml2021_oppo}

\icmltitlerunning{On the Memorization Properties of Contrastive Learning}

\begin{document}

\twocolumn[
\icmltitle{On the Memorization Properties of Contrastive Learning}




\begin{icmlauthorlist}
\icmlauthor{Ildus Sadrtdinov}{hse}
\icmlauthor{Nadezhda Chirkova}{hse}
\icmlauthor{Ekaterina Lobacheva}{hse}
\end{icmlauthorlist}

\icmlaffiliation{hse}{HSE University}

\icmlcorrespondingauthor{Ildus Sadrtdinov}{isadrtdinov@hse.ru}
\icmlcorrespondingauthor{Ekaterina Lobacheva}{elobacheva@hse.ru}

\icmlkeywords{Machine Learning, ICML}

\vskip 0.3in
]



\printAffiliationsAndNotice{} % otherwise use the standard text.

\begin{abstract}
Memorization studies of deep neural networks (DNNs) help to understand what patterns and how do DNNs learn, and motivate improvements to DNN training approaches. 
In this work, we investigate the memorization properties of SimCLR, a widely used contrastive self-supervised learning approach, and compare them to the memorization of supervised learning and random labels training. 
We find that both training objects and augmentations may have different complexity in the sense of how SimCLR learns them. 
Moreover, we show that SimCLR is similar to random labels training in terms of the distribution of training objects complexity.
\end{abstract}

\section{Introduction}

Modern deep neural networks (DNNs) are often trained in a supervised learning setup, which requires a significant amount of labeled data. However, data labeling is often time-consuming and costly, as it involves human expertise. Thus, it is common for computer vision to pretrain DNNs on some large labeled dataset, e.\,g.\,ImageNet~\cite{imagenet}, and then to fine-tune the model to a specific downstream task. The self-supervised learning paradigm provides a human labeling-free alternative to the supervised pretraining: recently developed contrastive self-supervised methods show results, comparable to
ImageNet pretraining on a range of downstream tasks~\cite{simclr, moco}. In this work, we focus on SimCLR~\cite{simclr}, a commonly-used contrastive pretraining approach that applies two strong augmentations to each image in a mini-batch and asks the network to find correct pairing.

\begin{figure}[t]
\begin{center}
\centerline{\includegraphics[width=\columnwidth]{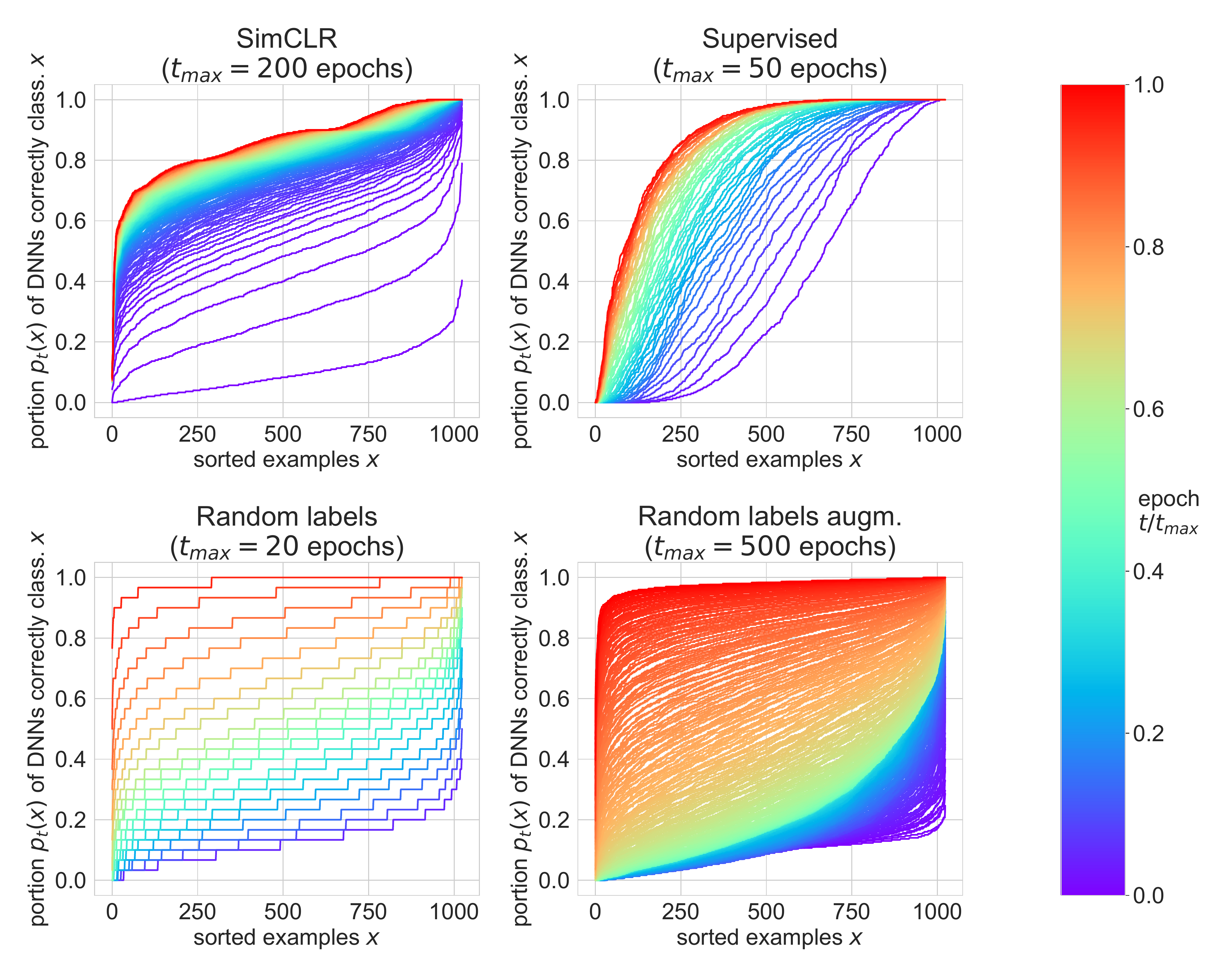}}
\caption{
Memorization profiles of ResNet-18
for a subset of 1K CIFAR-10 training examples. The curves $p_t(x)$ for different epochs $t$ are sorted independently, $t_{\max}$ denotes a maximum number of training epochs. The curves for random labels have staircase-like shape, as there is no averaging across augmentations. 
}
\label{training-dynamics}
\end{center}
\vskip -0.2in
\end{figure}

\begin{figure*}[t]
\begin{center}
\centerline{\begin{tabular}{cp{0.3cm}c}
{\small Averaged across augmentations}& &{\small Fixed augmentations}\\
  \includegraphics[width=0.45\textwidth]{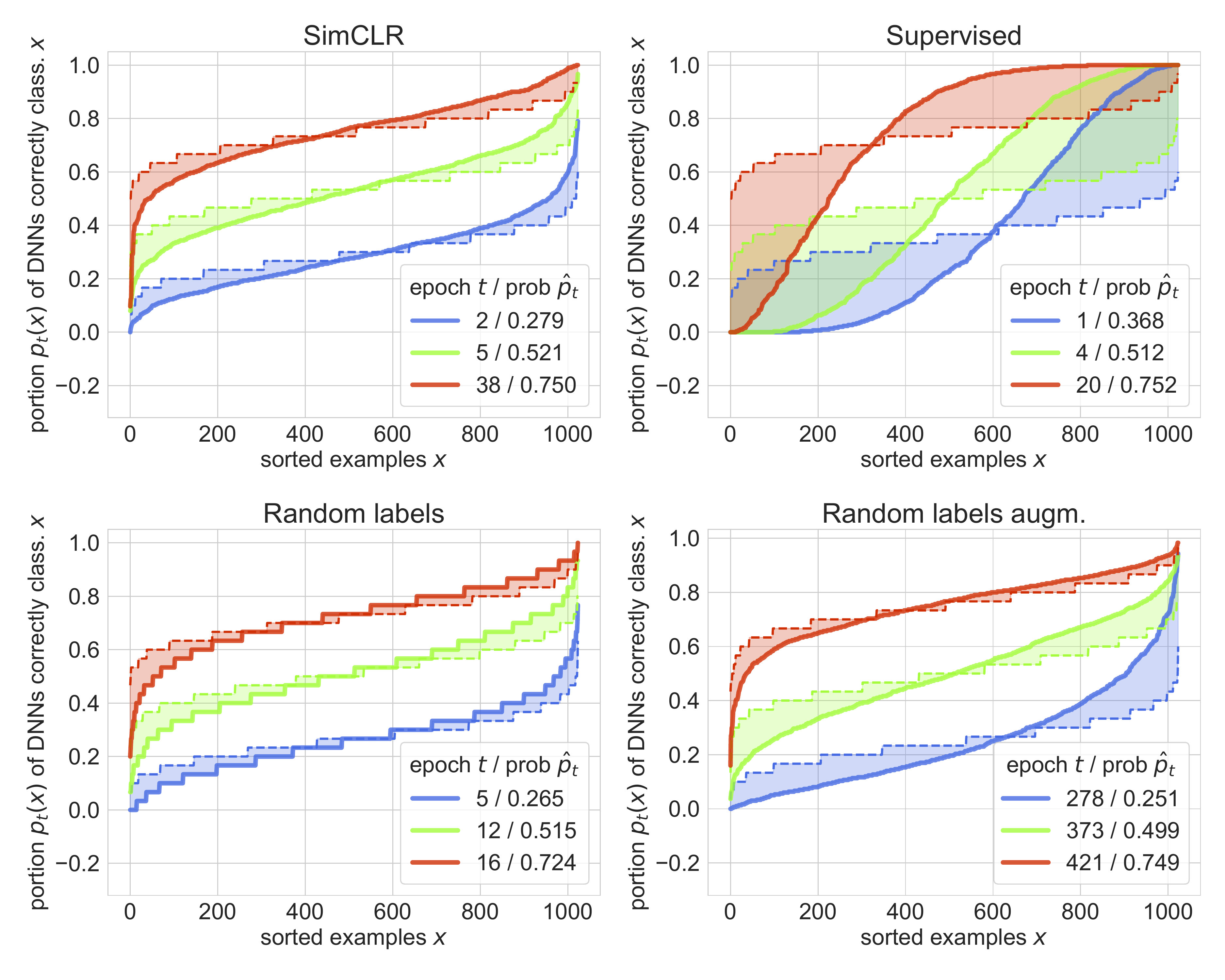}& &\includegraphics[width=0.45\textwidth]{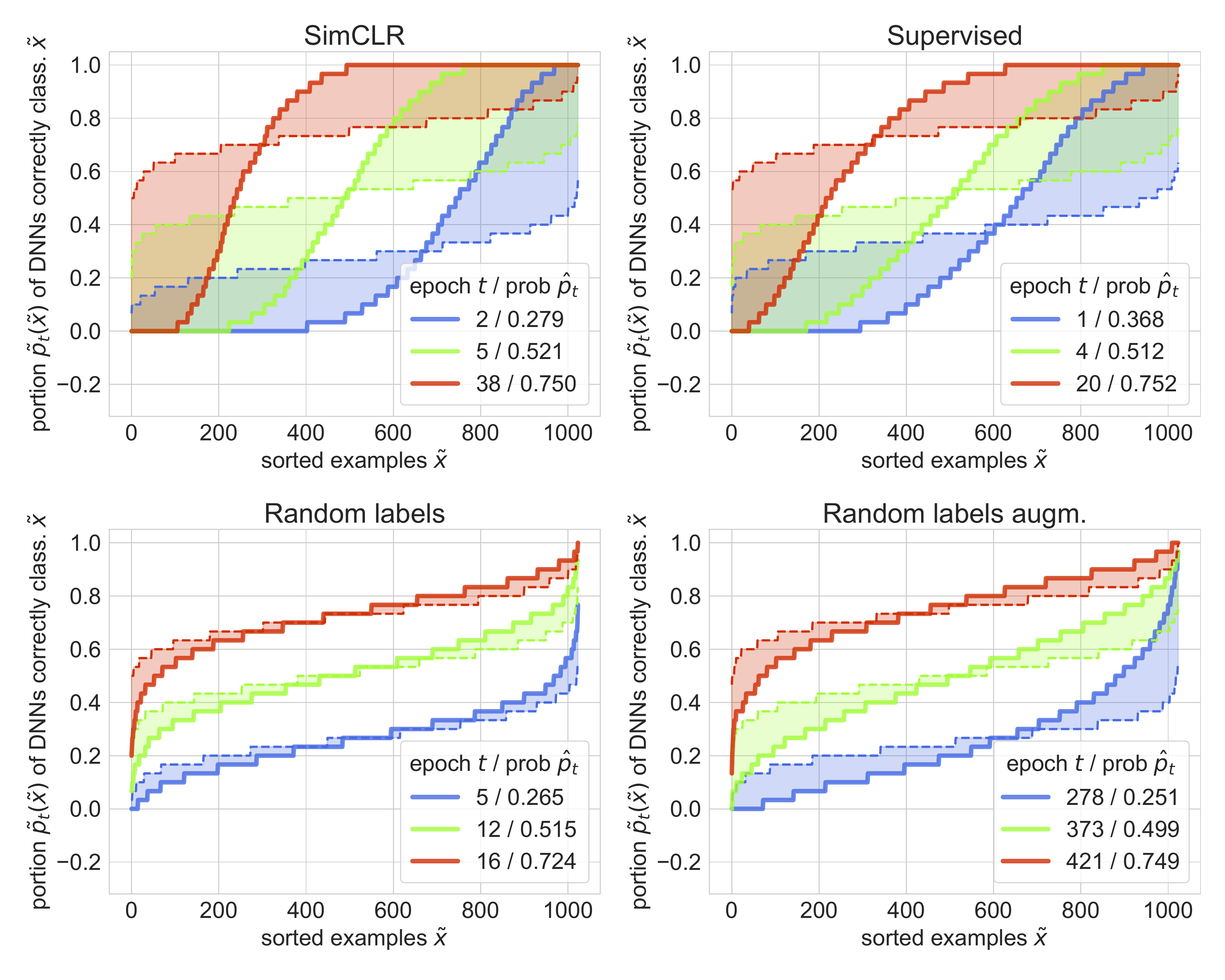}
  \end{tabular}
  }
\caption{Comparing memorization profiles to binomial noise curves. The graphs on the left and the right correspond to averaging across augmentations and fixed augmentations, respectively. The curves of $p_t(x)$ and $\tilde{p}_t (\tilde{x})$ are plotted with solid lines, while sorted binomial noise is marked with dotted lines. Here $\hat{p}_t$ denotes a maximum likelihood estimation for binomial sampling.} 
\label{binom-noise}
\end{center}
\vskip -0.2in
\end{figure*}

Despite DNNs achieve high results in various applications, the reasons for their prominent generalization are still not well understood. One of the powerful tools for analysing generalization properties are \emph{memorization studies} which investigate what patterns and how do DNNs learn and motivate improvements to DNN training approaches. 
A pioneer work of \citet{randlabels} showed that the capacity of modern DNNs is sufficient to fit perfectly even randomly labeled data. According to classic learning theory, such a huge capacity should lead to catastrophic overfitting, however, recent works~\cite{doubledescent} show that in practice increasing DNN capacity further improves generalization. \citet{freqdomain} suggest the reason for DNNs success is that they are implicitly biased towards learning low-frequency components, i.\,e.\,meaningful dependencies in the data, and learn high-frequency components, i.\,e.\,noise, only after learning low-frequency components. These works motivate the use of early stopping for DNN training. The phenomena of learning ``easier'' patterns before ``complex'' patterns is also supported in the work of~\citet{memorization} which shows that during supervised DNN training on image classification task, some images are almost always learned at the very beginning of training while other images are usually learned only at the end of training. This work suggests that some training images are easier to learn for the DNN than others and motivates the more careful training object ordering when training DNNs~\cite{mentornet}. 

The memorization properties has been widely studied for the supervised setup, while little is known about how self-supervised models memorize. In this work, we aim at closing this gap and investigate the memorization properties of the SimCLR approach, relying on the methodology of~\citet{memorization}. We firstly verify the hypothesis that some training images may be easier to learn than others while training DNN in the self-supervised manner. We find that although the described effect holds, the contrast between different objects' complexities is much less for self-supervised learning than for the supervised one and is similar to training with random labels. Moreover, inspired by the important role of data augmentations in contrastive pretraining, we conduct another set of experiments regarding the memorization of image--augmentation pairs. We find that data augmentations indeed have different complexities for SimCLR and some of them are much easier to learn for the DNN than others. Our results may motivate future studies on improving contrastive pretraining approaches by advanced augmentation ordering.

\section{Background: Self-Supervised Learning}
\label{sec:bcgd}
The idea of self-supervised pretraining is to construct a synthetic training task complex enough to force a neural network to learn semantic features. Early works in this field propose various image-level pretext tasks like jigsaw puzzle solving~\cite{jigsaw}, image colorization~\cite{colorization}, and rotation angle prediction~\cite{rotation}. \citet{surrogate} construct the pretext task by treating each training image as an independent ``class'' and generating ``instances'' of this ``class'' by applying various data augmentations. More recent studies~\cite{simclr, moco} rely on contrastive losses, that compel the neural network to discriminate views of different images. As contrastive pretraining approaches achieve higher results than pretext task-based approaches, we focus on the former group and select a commonly used approach from this group, SimCLR~\cite{simclr}.  This method generates two random augmentations as views for each image in the mini-batch, and the objective of the training procedure is to bring closer representations of the same image while moving apart the representations of distinct images.
That is, given a mini-batch of $B$ images, the SimCLR training objective is equivalent to the classification task inside the mini-batch with $2B-1$ classes: the correct class for each augmented view of an image is a paired view of the same image. The scheme of the SimCLR approach is given in Fig.~\ref{simclr} in Appendix. In the rest of the work, self-supervised learning denotes contrastive learning and specifically SimCLR approach.

\section{Methodology}

Our approach to studying memorization and training dynamics is inspired by the work of~\citet{memorization}. Although the approach was initially proposed for supervised learning, it can be easily adapted to self-supervised learning if we cast it as a classification problem, see Section~\ref{sec:bcgd}. We train $N=30$ DNNs, all having the same architecture but different random initialization, and for a given image $x$ and epoch $t$, we estimate the portion $p_t(x)$ of DNNs that correctly recognize the image at  epoch $t$. For classification, recognition means correct classification, and for SimCLR, recognition means finding the correct pair for the image. In all experiments comprising data augmentation, $p_t(x)$ also includes averaging across $M=10$  random augmentations. 

An image is called \emph{easy-to-memorize} if it is correctly recognized by almost all networks at \emph{early} epochs, and \emph{hard-to-memorize}~--- if this happens only at \emph{late} epochs or does not happen at all. 
For each epoch $t$, we can visualize its \emph{memorization profile} by sorting $p_t(x)$ over images $x$ in data subset $X$.  If there is no distinction between easy- and hard-to-memorize images, the memorization profile  at any epoch $t$ can be closely approximated with binomial noise curve, see more details in Section~\ref{sec:exp1}. Example evolution of memorization profiles over epochs is presented in Fig.~\ref{training-dynamics} and a more detailed description of profiles is given in Appendix~\ref{appendix-B}. 

We conduct experiments on the CIFAR-10 dataset~\cite{cifar} and consider three learning paradigms: supervised learning, self-supervised learning and random labels training. 
Random labels training is usually performed without augmentation, however, augmentations are crucial for SimCLR. Thus, we consider two options for random labelling training: training with and without augmentations.
SimCLR utilizes more intensive augmentations, while supervised and random labels setups use weaker augmentations common for the CIFAR-10 dataset (see Appendix~\ref{appendix-A} for more details).

In our experiments, we use ResNet-18 architecture~\cite{resnet}. We use Adam~\cite{adam} to train SimCLR models and stochastic gradient descent (SGD) with momentum $m=0.9$~--- for other setups. All methods use cosine annealing learning rate schedule~\cite{cosine}. The training hyperparameters are set as follows: initial learning rate $\eta=3\cdot 10^{-4}$, weight decay $\lambda=10^{-4}$, and batch size $B=256$. Softmax temperature is $\tau=0.07$ for SimCLR and $\tau=1$ for other methods. The output dimensionality of SimCLR is $d=128$.

\section{Experiments}

\subsection{Comparison to Binomial Noise}
\label{sec:exp1}
We begin with investigating whether there is a distinction between easy- and hard-to-memorize images in self-supervised training (as well as in other considered paradigms), or all images have the same complexities. 
In the latter case, $p_t(x)$ would be independent of $x$ and the memorization profile at any epoch could be closely approximated with binomial noise curve. 
Therefore, to answer the stated question, we can estimate the parameter $p$ on $\{{p}_t (x)\}_{x \in X}$, sample the corresponding binomial noise, and compare the sorted binomial noise curve to the curve of  sorted ${p}_t (x)$ (memorization profile). 

\begin{figure}[]
\begin{center}
\centerline{
\begin{tabular}{c}
{\small Averaged across augmentations}\\
  \includegraphics[width=\columnwidth]{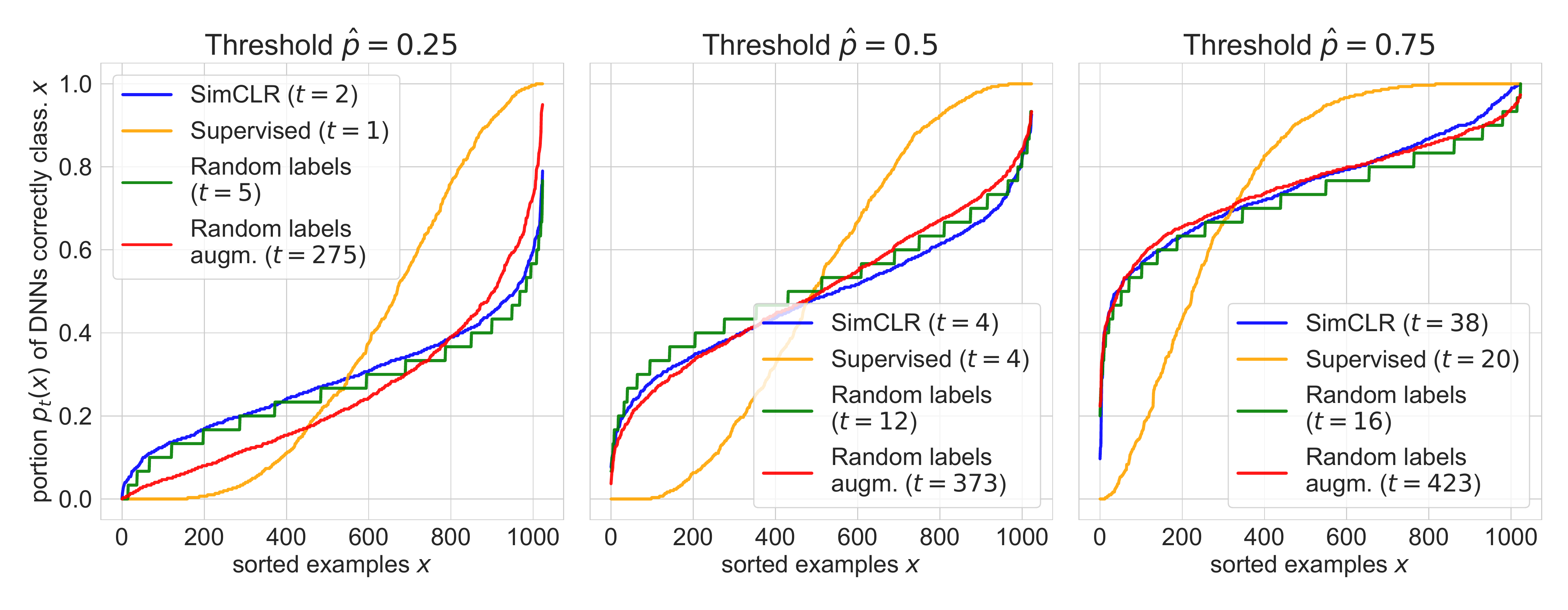}\\
  {\small Fixed augmentations}\\
  \includegraphics[width=\columnwidth]{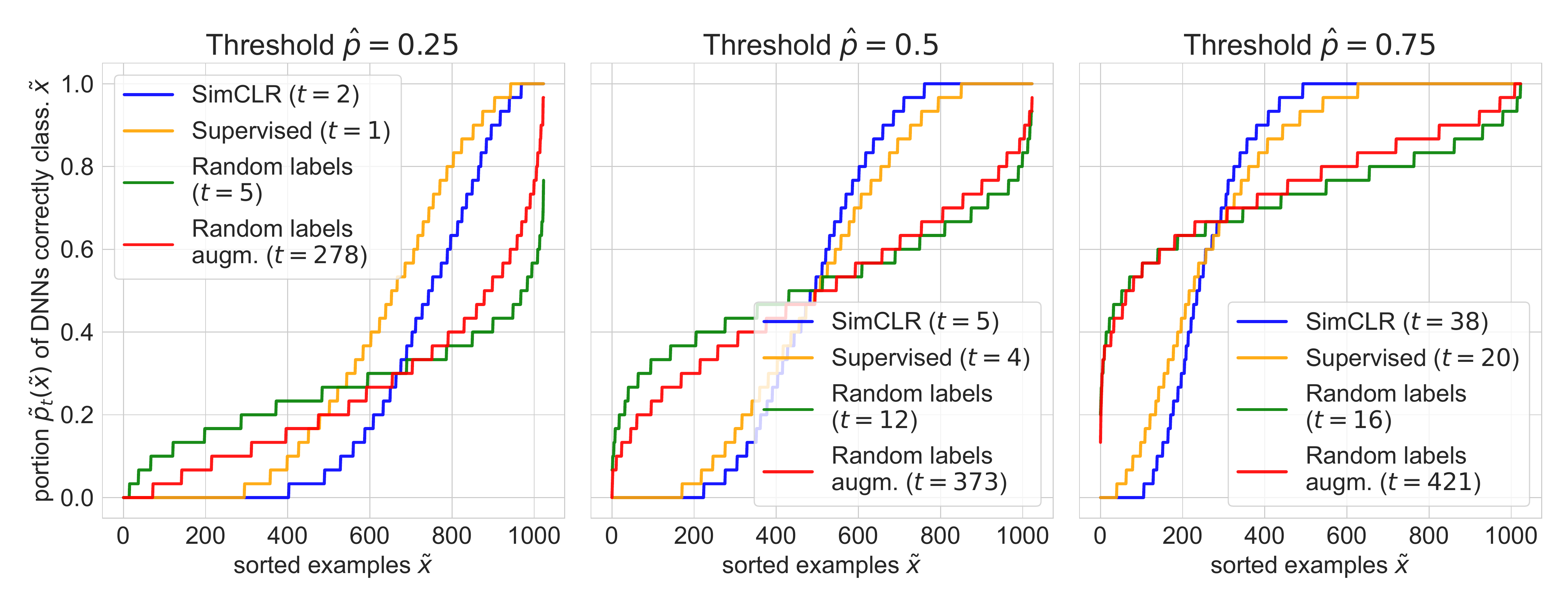}\\
  \end{tabular}
  }
\caption{Memorization profiles comparison. Notation in brackets shows the number of training epochs $t$ for each method.}
\label{equal-areas}
\end{center}
\vskip -0.2in
\end{figure}

Fig.~\ref{binom-noise} (left) presents the results of comparing memorization profiles to binomial curves at different epochs, for all considered learning paradigms. 
According to the plots, the memorization profiles differ from the binomial curves in all cases, indicating the presence of easy- and hard-to-learn training images in all considered setups, including self-supervised learning. 
However, the difference between memorization profiles and their approximations is much larger for supervised learning than for other paradigms. In other words, \emph{the contrast in complexity between easy- and hard-to-learn images is much higher for supervised learning than for the self-supervised one.} 

Further, motivated by the important role of data augmentations in self-supervised learning, we conduct a similar experiment but considering memorization profiles w.\,r.\,t. image--augmentation pairs, in contrast to the previous experiment, where we considered profiles w.\,r.\,t. images itself. Particularly, in the previous experiment, $p_t(x)$ included averaging across $M=10$ random augmentations, and in this experiment,
we fix one augmentation\footnote{In SimCLR, we fix two augmentations, required by method.} for each image $x$, obtaining an image-augmentation pair $\tilde x$, and construct memorization profiles $\tilde p_t(\tilde x)$ for the set of these pairs.

Fig.~\ref{binom-noise} (right) presents the results of comparing memorization profiles to binomial curves at different epochs, with fixed augmentations. 
Notably, in contrast to the previous experiment, the SimCLR memorization profiles now substantially differ from their binomial approximations.
That is, \emph{in contrastive learning, some image--augmentation pairs are much easier to learn than others}. 
For supervised learning and random labels experiments, data augmentation is not a part of problem statement, and fixing augmentations does not change results significantly. In Appendix~\ref{appendix-B}, we explain that comparing to the binomial approximation is even more theoretically grounded for the fixed augmentations case than for the case of averaging across augmentations.

\begin{figure}[]
\begin{center}
\centerline{\includegraphics[width=\columnwidth]{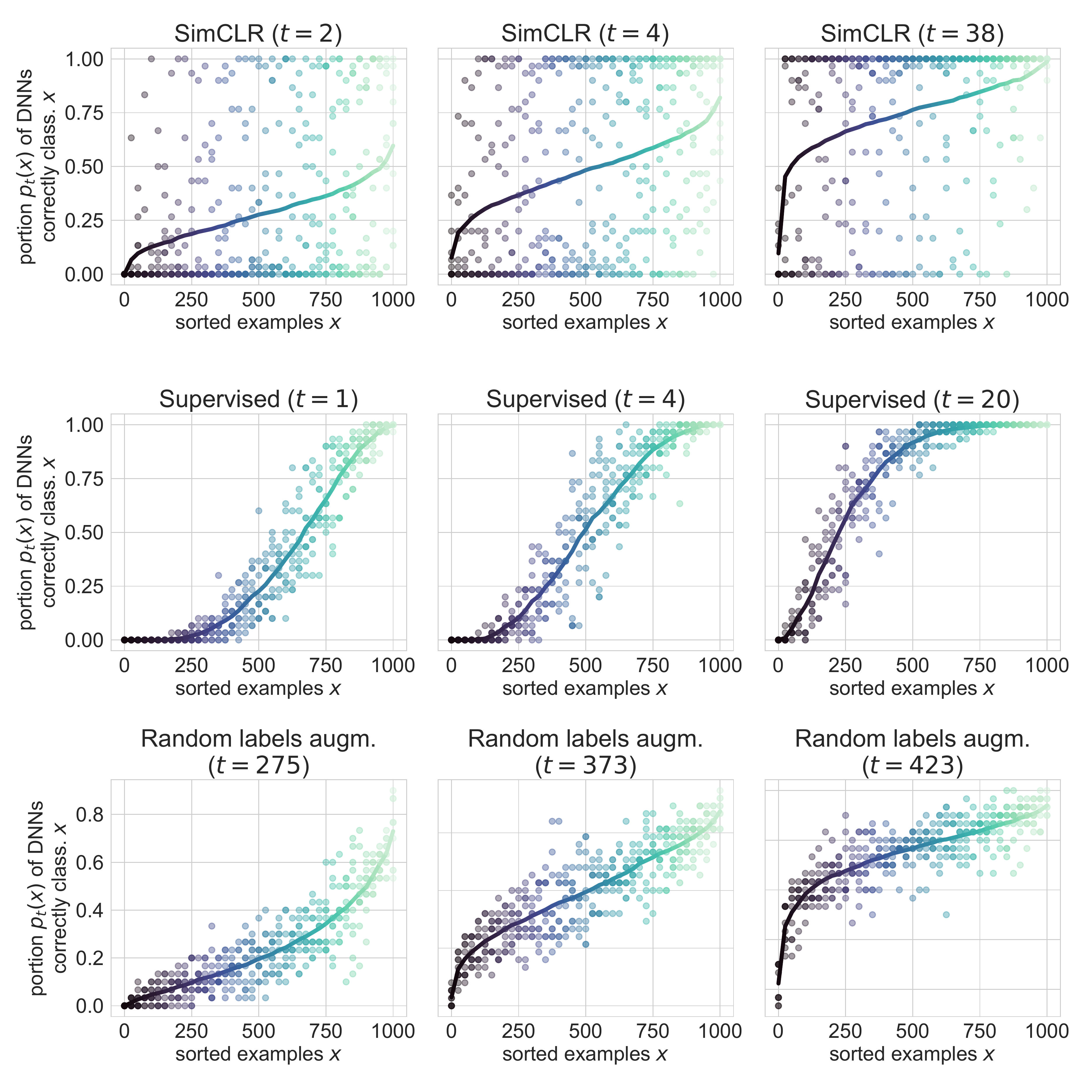}}
\caption{Augmentations' impact on complexity. Each point represents the probability $\tilde{p}_t(\tilde{x})$ for a single image--augmentation pair $\tilde{x}$. Solid lines correspond to the average values across the augmentations, i.\,e.\,$p_t(x)$. Every 25th example is plotted.}
\label{augments}
\end{center}
\vskip -0.2in
\end{figure}

In Fig.~\ref{augments}, we plot the probability $\tilde{p}_t(\tilde{x})$ for $M=10$ augmented versions of each image. For the supervised and random labels setups, the values of $\tilde{p}_t(\tilde{x})$ have relatively low variance, concentrating near the average value $p_t(x)$. In contrast, most of the SimCLR views $\tilde{x}$ are either too easy or too complex to be memorized, providing one more evidence that some image--augmentation pairs are much more complex than the others.

\subsection{Comparison of learning paradigms}

We now compare memorization profiles between different learning paradigms. 
Fig.~\ref{training-dynamics} presents the dynamics of how memorization profiles (averaged across augmentations) evolve over epochs, for all considered learning paradigms. Each line in these plots denotes the memorization profile at some training epoch.
On the one hand, in terms of the density of the plots, there is a clear distinction between methods that learn meaningful data representations, namely supervised and self-supervised methods, and the random labels experiment. Particularly, for both supervised and self-supervised approaches, the memorization profiles quickly concentrate in the top part of the plot and continue slowly evolving there, while for random labels, the profile uniformly evolves from bottom to top. On the other hand, in terms of the lines' shape, the memorization profiles of self-supervised learning look similar to random labels' ones.
Below we look at this effect closer.
We also note that two variants of random labels training behave similarly.

To compare the shape of the memorization profiles, we need to align their epochs.
It is natural to align the methods by the mean value of $p_t(x)$~/~$\tilde p_t(\tilde x)$, which is equal to the area under sorted curve of $p_t(x)$~/~$\tilde p_t(\tilde x)$ (area under memorization profile). According to Fig. \ref{training-dynamics}, the area increases with time, corresponding to the evolution of training phases. 
We consider several thresholds $\hat{p} \in \{0.25, 0.5, 0.75\}$ and select a training epoch $t$ with the mean value of $p_t(x)$ closest to  each $\hat{p}$. The resulting comparison of the memorization profiles of different learning paradigms is presented in Fig. \ref{equal-areas}, for both averaging over augmentations (top row) and fixed augmentation (bottom row) scenarios. With averaging over augmentations, the supervised setup differs significantly from all other setups: the sorted curves follow a characteristic S-shape, reflecting the presence of extremely easy- and hard-to-memorize training images. The memorization profiles of the rest of the methods are quite similar to each other and indicate much less contrast in the images' complexities than in supervised learning. However, with fixed augmentations, SimCLR's memorization profiles become similar to the supervised learning's ones. 

In the described sense of comparing memorization profiles w.\,r.\,t. training images (not image-augmentation pairs), \emph{the training dynamics of SimCLR is similar to random labels training dynamics}. The possible explanation is that in SimCLR, each training image actually represents its own ``class'' that the DNN has to remember. At the same time, in the random labels experiment, the DNN also has to memorize a (random) class for each training image.

In Appendix~\ref{appendix-C}, we present additional comparison of representations learned by four considered training setups.

\section{Conclusion}
In this work, we investigated the memorization properties of SimCLR and found that (1) some training \emph{images} are easier to learn than others in the self-supervised paradigm as well as in the supervised one, but the contrast between images' complexities is much higher for supervised learning; (2) some \emph{augmented images} are much easier to learn than others in self-supervised learning, with high contrast between them; (3) in the sense of comparing memorization profiles w. r. t. training images, the training dynamics of SimCLR is similar to random labels training dynamics. Our findings may motivate the future development of advanced data augmentation ordering in self-supervised approaches, e.\,g.\,progressive increasing of augmentation strength.

\section*{Acknowledgments}
The work was supported by the Russian Science Foundation grant \textnumero 19-71-30020. This research was supported in part through computational resources of HPC facilities at NRU HSE.

\bibliography{example_paper}
\bibliographystyle{icml2021_oppo}

\appendix

\section{Augmentations}
\label{appendix-A}

\begin{figure}[t]
\begin{center}
\begin{tikzpicture}
\node[inner sep=0pt] (cow0) at (-1, 0)
    {\includegraphics[width=.25\columnwidth]{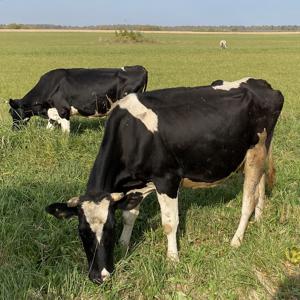}};
\node[inner sep=0pt] (cow1) at (2.25, 1.25)
    {\includegraphics[width=.25\columnwidth]{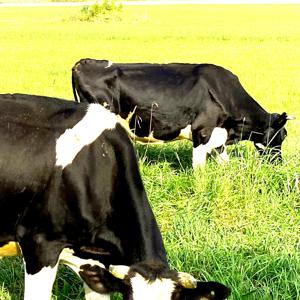}};
\node[inner sep=0pt] (cow2) at (2.25, -1.25)
    {\includegraphics[width=.25\columnwidth]{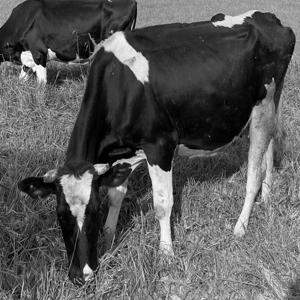}};
\node[inner sep=3pt] (z1) at (5.5, 1.25) {\large $z_{2i-1}$};
\node[inner sep=3pt] (z2) at (5.5, -1.25) {\large $z_{2i}$};
\node[inner sep=0pt] (f1) at (4.25, 1.7) {\large $f_{\theta}(\cdot)$};
\node[inner sep=0pt] (f2) at (4.25, -1.7) {\large $f_{\theta}(\cdot)$};
\node[inner sep=0pt] (x0) at (-1, 1.3) {\large $x_i$};
\node[inner sep=0pt] (x1) at (2.25, 2.55)
    {\large $\tilde{x}_{2i-1} \sim \mathcal{A}(\tilde{x}|x_i)$};
\node[inner sep=0pt] (x2) at (2.25, -2.55)
    {\large $\tilde{x}_{2i} \sim \mathcal{A}(\tilde{x}|x_i)$};

\draw[-latex, thick] (cow0.east) -- (cow1.west);
\draw[-latex, thick] (cow0.east) -- (cow2.west);
\draw[-latex, thick] (cow1.east) -- (z1.west);
\draw[-latex, thick] (cow2.east) -- (z2.west);
\draw[latex-latex, thick] (z1.south) -- (z2.north);
\end{tikzpicture}

\caption{Scheme of the SimCLR algorithm. Here $\mathcal{A}(\tilde{x}|x)$ denotes the distribution of augmentations for image $x$. Given a mini-batch of images $\{x_i\}_{i=1}^B$, the method samples two views $\tilde{x}_{2i-1}, \tilde{x}_{2i} \sim \mathcal{A}(\tilde{x}|x_i)$ for each image $x_i$. The views are passed through the DNN $f_{\theta}$, generating a doubled mini-batch of representations $\{z_j\}_{j=1}^{2B}$. The DNN is trained to maximize the similarity between the corresponding representations, i.\,e.\,$z_{2i-1}$ and $z_{2i}$, while the distinct representations are pushed away.}
\label{simclr}
\end{center}
\vskip -0.2in
\end{figure}
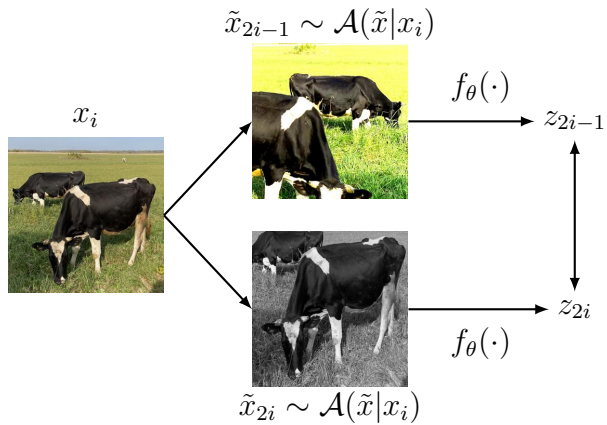

We use SimCLR augmentations as described in the original paper \cite{simclr}, including random crop, horizontal flipping, color jitter, grayscale conversion, and gaussian blur. In contrast, supervised learning and random labels training use random crop, horizontal flipping, and per-channel normalization with a vector of means $\mu=(0.4914, 0.4822, 0.4465)$ and a vector of standard deviations $\sigma=(0.2023, 0.1994, 0.2010)$, specific for the CIFAR-10 dataset. The scheme of data-processing pipeline of SimCLR is presented in Fig.~\ref{simclr}.

\section{Binomial Noise}
\label{appendix-B}
In this section, we give a more detailed description of memorization profiles and provide the reasoning for approximating them with Binomial curves, in case all images~/~image--augmentation pairs have same memorization complexities.

Let $\mathcal{T}(\theta|t)$ denote the distribution of DNN weights $\theta$ after $t$ epochs of training (including random initialization, SGD objects shuffling and augmentations sampling). Let $\mathcal{A}(\tilde{x}|x)$ be the distribution of augmentations $\tilde{x}$ for image $x$. Let $I_{\theta}(x)$ be an indicator of correct classification, i.\,e.\,it equals $1$ if DNN with weights $\theta$ classifies image $x$ correctly and $0$ otherwise. We consider the probability $p_t(x)$ of correct classification over the distribution of weights and augmentations:
\begin{equation}
\begin{gathered}
    p_t(x) = \mathbb{P}_{\substack{\theta \sim \mathcal{T}(\theta|t) \\ \tilde{x} \sim \mathcal{A}(\tilde{x}|x)}} \Big(I_{\theta}(\tilde x)=1\Big) \approx \\
    \approx \frac{1}{NM} \sum_{i=1}^N \sum_{j=1}^M I_{\theta_i} (\tilde{x}_j)
\end{gathered}
\end{equation}
We estimate $p_t(x)$ via Monte-Carlo method by sampling $M=10$ augmentations $\tilde{x}_j \sim \mathcal{A}(\tilde{x}|x)$ for each image $x$ and training $N=30$ DNNs with weights $\theta_i \sim \mathcal{T}(\theta|t)$ from different random initializations. The distribution of augmentations $\mathcal{A}(\tilde{x}|x)$ is the same as used during training. In particular, one of the random labels setups uses degenerate distribution $\mathcal{A}(\tilde{x}|x) = \delta(\tilde{x}-x)$. The \textit{memorization profiles} are obtained by sorting $p_t(x)$ over $x \in X$, where $X$ is a subset of training images.

We also consider fixing one augmentation $\tilde{x} \sim \mathcal{A}(\tilde{x}|x)$ for each training image $x$ and estimating the probability $\tilde{p}_t (\tilde{x})$ of correct classification over the distribution of weights:
\begin{equation}
\begin{gathered}
    \tilde{p}_t (\tilde{x}) = \mathbb{P}_{\theta \sim \mathcal{T}(\theta|t)} \Big(I_{\theta} (\tilde{x}) = 1 \Big) \approx \frac{1}{N} \sum_{i=1}^N I_{\theta_i} (\tilde{x})
\end{gathered}
\end{equation}
The probability $\tilde{p}_t (\tilde{x})$ differs from $p_t (x)$ because there is no averaging across augmentations. The memorization profiles in this case are obtained by sorting $\tilde p(\tilde x)$ over $\tilde x$.

Let us now suppose that all image--augmentation pairs have equal memorization complexities. Then, the random variable $I_{\theta}(\tilde x)$ depends on $\theta \sim \mathcal{T}(\theta|t)$ but not on $\tilde x$. As this indicator is binary, we can assume it has the Bernoulli distribution with the probability of success $p$ independent from $\tilde x$. The indicator depends only on $\theta$, thus if $\theta_1, \theta_2 \sim \mathcal{T}(\theta|t)$ are independent, then the corresponding indicators $I_{\theta_1}(x), I_{\theta_2}(x)$ are independent as well. Now the random variable $\tilde{p}_t (\tilde{x})$ can be expressed as a sum of independent Bernoulli variables with the same probability of success $p$, and this sum follows the binomial distribution (up to a constant $1/N$):
\begin{equation}
    \tilde{p}_t (\tilde{x}) \sim \frac{1}{N} \sum_{i=1}^N \text{Bern}(p) = \frac{1}{N} \text{Bin}(N, p) 
\end{equation}

However, the same theoretical derivation is not suitable for $p_t(x)$. The averaged across augmentations indicator follows some distribution over the set $\{0, 1/M, \dots, 1\}$, so we no longer can express $p_t(x)$ as a sum of Bernoulli random variables:
\begin{equation}
    p_t(x) \approx \frac{1}{N} \sum_{i=1}^N \left(\frac{1}{M} \sum_{j=1}^M I_{\theta_i} (\tilde{x}_j) \right)
\end{equation}
This discrete distribution is complex for theoretical analysis, depending on the distribution of augmentations $\mathcal{A}(\tilde{x}|x)$.

Overall, there are theoretical foundations to compare the memorization profiles of $\tilde{p}_t(\tilde{x})$ to binomial samples. Although the same does not hold for $p_t(x)$, we can still compare the sorted curves, thinking of binomial noise as an approximation for equal images' complexities.

\section{Additional Experimental Results}
\label{appendix-C}

\subsection{t-SNE Visualization}

\begin{figure}[t]
\begin{center}
\centerline{\includegraphics[width=\columnwidth]{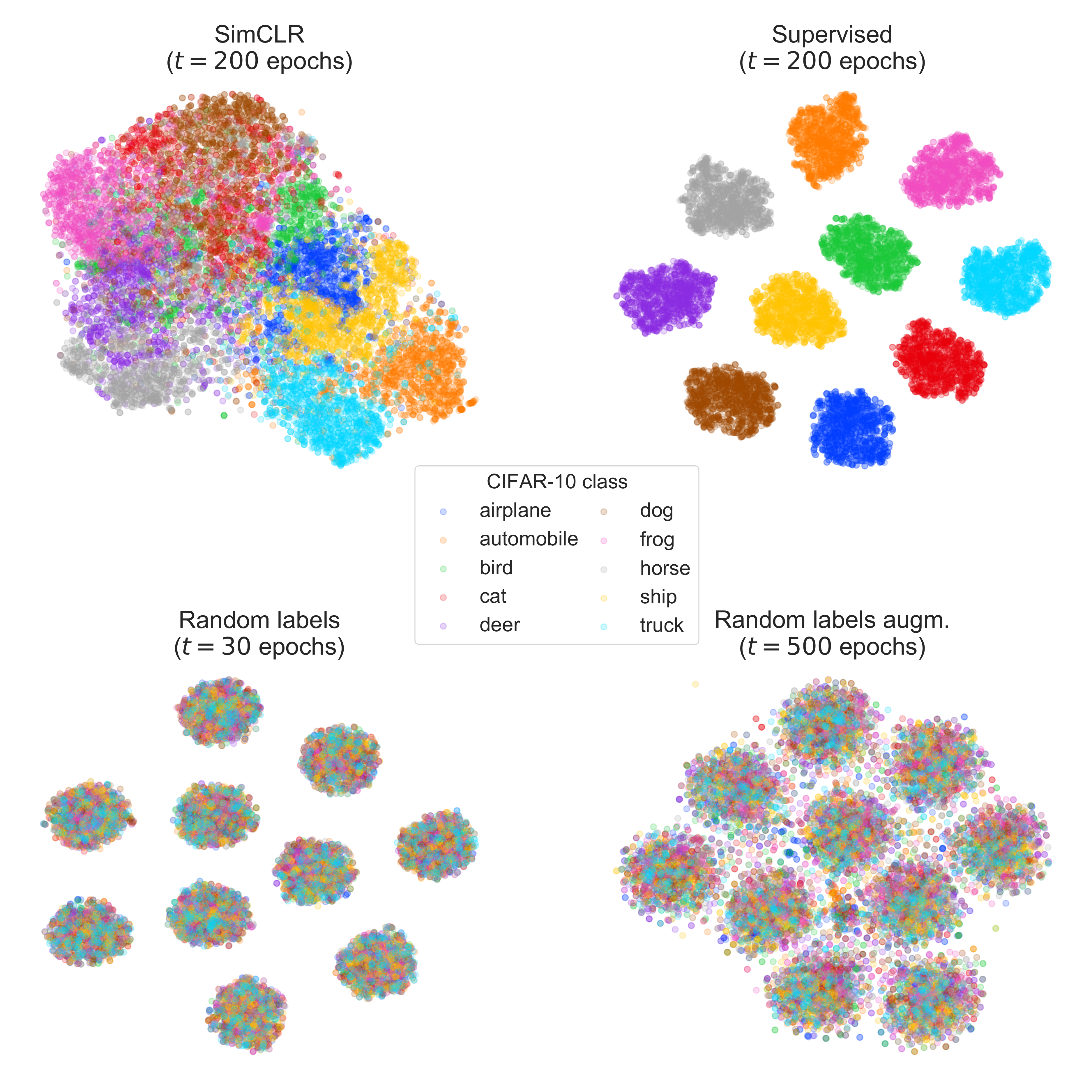}}
\caption{t-SNE visualization of learned representations of training images.} 
\label{tsne}
\end{center}
\vskip -0.2in
\end{figure}

To further compare the considered training methods, we analyze the representations of training images learned by  DNNs and visualize them with the t-SNE algorithm~\cite{tsne}. The result is shown in Fig.~\ref{tsne}. The dimensionality of latent space is $D=512$. Notably, the representations of classification tasks, namely the supervised and both random labels setups, form distinct clusters that correspond to the training objective. In contrast, SimCLR training forces the DNN to locate the representations more densely. Interestingly, the representations of each CIFAR-10 class concentrate close to each other, although the class labels are not available for SimCLR training. This semantic structure of the SimCLR latent space is evidence of high generalization of contrastive self-supervised methods.

\subsection{DNN Activations Visualization}

\begin{figure*}[t]
\begin{center}
\centerline{\includegraphics[width=\textwidth]{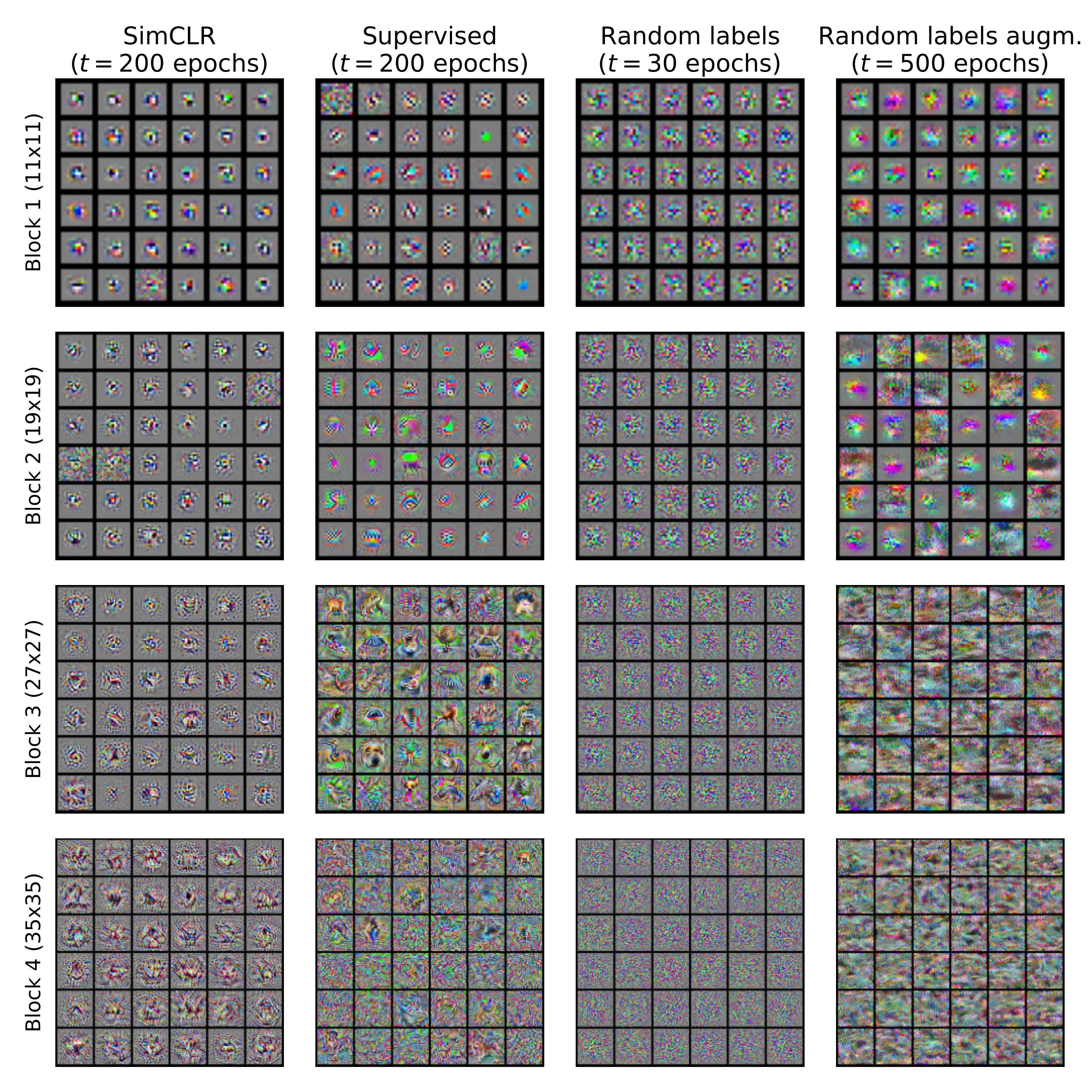}}
\caption{Trained DNNs activations vizualization. Each column matches a particular training setup, and each row represents activations from the same ResNet-18 block. Notation in brackets shows the receptive field size. The feature map channels were chosen randomly.} 
\label{activations}
\end{center}
\vskip -0.2in
\end{figure*}

To demonstrate the difference between the training paradigms, we visualize the activations of fitted convolutional DNNs using the approach of~\citet{visual}. This method retrieves an image that maximizes some specific activation via gradient ascent in pixel space. To visualize the generated images, we normalize each image to have zero mean and unit variance and pass the images through the element-wise logistic function $\sigma(y) = 1/(1 + e^{-y})$ to fit the RGB-range of $[0, 1]$. The results are presented in Fig.~\ref{activations}.

The images of random labels training look like uninterpretable noise, but adding augmentations to this setup smooths the visualizations. SimCLR and supervised learning images look similar, evolving from simple patterns on early layers to complex patterns on later layers. However, some attributes of the CIFAR-10 dataset, e.\,g.\,a horse head (Block 3, last column, second-to-last row), can be seen in the supervised images, while the SimCLR images seem to be less interpretable. This observation provides intuition for the difference between two setups: 
supervised learning aims at determining image attributes specific for each class presented in the train set. In contrast, a DNN trained with the SimCLR algorithm has to learn more general patterns from the images, which survive after intensive augmentations and allow to couple the views correctly.

\end{document}